\documentclass[sigconf,authorversion]{acmart}

\copyrightyear{2018} 
\acmYear{2018} 
\setcopyright{acmlicensed}
\acmConference[CIKM '18]{The 27th ACM International Conference on Information and Knowledge Management}{October 22--26, 2018}{Torino, Italy}
\acmBooktitle{The 27th ACM International Conference on Information and Knowledge Management (CIKM '18), October 22--26, 2018, Torino, Italy}
\acmPrice{15.00}
\acmDOI{10.1145/3269206.3271702}
\acmISBN{978-1-4503-6014-2/18/10}
\usepackage{booktabs} 

\usepackage{graphicx}
\usepackage{amssymb,amsmath,amsthm}
\usepackage{color}

\newtheoremstyle{mydef}%
{}{}{\normalfont}{}{\itshape}{.\,}{ }{}
\theoremstyle{mydef}
\newtheorem{definition}{Definition}

\newtheoremstyle{myprob}%
{}{}{\normalfont}{}{\scshape}{.\,}{ }{}
\theoremstyle{myprob}
\newtheorem{problem}{Problem}

\newtheoremstyle{mylayer}%
{}{}{\normalfont}{}{\itshape}{.\,}{ }{}
\theoremstyle{mylayer}
\newtheorem{layer}{}

\begin{document}

\title{Retrieve-and-Read: Multi-task Learning of Information Retrieval and Reading Comprehension}

\author{Kyosuke Nishida, Itsumi Saito, Atsushi Otsuka, Hisako Asano, and Junji Tomita}
\affiliation{%
  \institution{NTT Media Intelligence Laboratories, NTT Corporation}%
}
\email{nishida.kyosuke@lab.ntt.co.jp}

\begin{abstract}
This study considers the task of machine reading at scale (MRS) wherein, given a question, a system first performs the information retrieval (IR) task of finding relevant passages in a knowledge source and then carries out the reading comprehension (RC) task of extracting an answer span from the passages. Previous MRS studies, in which the IR component was trained without considering answer spans, struggled to accurately find a small number of relevant passages from a large set of passages. In this paper, we propose a simple and effective approach that incorporates the IR and RC tasks by using supervised multi-task learning in order that the IR component can be trained by considering answer spans. Experimental results on the standard benchmark, answering SQuAD questions using the full Wikipedia as the knowledge source, showed that our model achieved state-of-the-art performance. Moreover, we thoroughly evaluated the individual contributions of our model components with our new Japanese dataset and SQuAD. The results showed significant improvements in the IR task and provided a new perspective on IR for RC: it is effective to teach which part of the passage answers the question rather than to give only a relevance score to the whole passage.
\end{abstract}



\maketitle

\section{Introduction}
Creating an AI capable of answering questions as well as people can has been a long-standing research problem. Recently, \textit{reading comprehension} (RC), a challenge to read a passage and then answer questions about it, has received much attention. Large and high-quality datasets that are sufficient to train deep neural networks have been constructed; in particular, the SQuAD dataset~\cite{RajpurkarZLL16} has brought significant progress such that the RC performance of AI is now comparable to that of humans.

In the SQuAD 1.1 dataset, each question refers to one passage of an article, and the corresponding answer is guaranteed to be a span in that passage
\footnote{The experiments reported in this paper used the SQuAD 1.1 dataset. SQuAD 2.0, which was recently released, additionally contains unanswerable questions based on one passage~\cite{RajpurkarJL18}. }. %
Thus, most of the current top-performing RC methods such as BiDAF~\cite{SeoKFH17} and QANet~\cite{Yu18} assume that one \textit{relevant} passage, which contains all the facts required to answer the question, is given when answering the question. 

We tackle the task of \textit{machine reading at scale} (MRS) wherein, given a question, a system retrieves passages relevant to the question from a corpus and then extracts the answer span from the retrieved passages. \citeauthor{ChenFWB17} proposed DrQA, which is
an open-domain QA system using Wikipedia's texts as a knowledge source by simply combining an exact-matching IR method with an RC method based on a neural network~\cite{ChenFWB17}. Their system showed promising results; however, the results indicated that the IR method, which retrieved the top five passages from five million articles for each question, was a bottleneck in terms of accuracy. It can retrieve passages that contain question words, but such passages are not always relevant to the question.

\begin{figure}[t!]
\centering
\includegraphics[width=.45\textwidth]{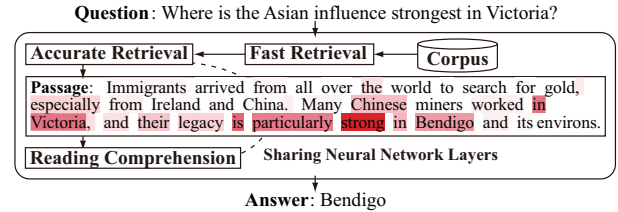}
\caption{Our machine-reading-at-scale system uses a corpus of passages as a knowledge source. Our neural network learns IR and RC tasks jointly. Its IR component accurately re-ranks the passages retrieved by using fast IR methods.}
\label{fig:concept}
\end{figure}

Here, we focus on the strong relationship between IR and RC. The RC capability of identifying an answer span in a passage will improve the IR capability of distinguishing between relevant and irrelevant passages. However, applying a model trained in RC to IR is not promising because the RC model, trained with only relevant passages, cannot indicate that there is no answer in irrelevant passages. We need to train a model so that it has both capabilities.

Recently, a joint neural model of IR and RC components, trained with reinforcement learning, was examined~\cite{WangAAAI2018}. It outperformed 
DrQA; however, the IR component still was a bottleneck. The IR component was indirectly trained with a \textit{distant supervision} reward, which indicates how well the answer string extracted by its RC component matches the ground-truth. We conjecture that this reward, which does not consider the answer span, may prevent the IR component from carefully considering the context of passages. 

Our main research goal is to investigate the impact of learning from answer \textit{spans} in IR for RC. For this, we propose a neural model that incorporates the IR and RC tasks by using supervised multi-task learning (MTL). It shares the hidden layers between IR and RC tasks and minimizes the joint loss of relevance scores in IR and answer spans in RC. Our model can be trained using standard RC datasets, such as SQuAD, consisting of the triples of a question, a passage, and an answer span. We use the triples in the datasets as positive (i.e., relevant) examples and generate negative examples from the datasets for the learning of the IR component.

Although our neural model can alleviate the bottleneck of IR accuracy, adapting it to the whole of a large-scale corpus causes computational complexity problems. We therefore introduce telescoping settings~\cite{MatveevaBBLW06}, where our IR model component re-ranks the outputs of fast exact-matching models that focus on eliminating higher irrelevant passages (Figure~\ref{fig:concept}). This idea enables our model to perform at a practical speed without loss of generality.

The main contributions of this study are as follows.

\begin{itemize}
\item We developed a \textit{Retrieve-and-Read} model for supervised MTL of IR and RC tasks that shares its hidden layers between the two tasks and minimizes the joint loss.

\item Our model with a telescoping setting exceeded the state-of-the-art by a significant margin on a large-scale MRS task, answering SQuAD questions using the full Wikipedia as the knowledge source.

\item We created a new dataset, Jp-News, which is based on Japanese news articles. This dataset is more difficult for IR models than SQuAD because of the existence of similar passages and articles on the same topics.

\item We thoroughly evaluated the effectiveness of MTL by investigating the individual contributions of our model components. We confirmed significant improvements in IR by learning from answer spans.
\end{itemize}

\section{Problem statement}
Let us state the problem that this study addresses.

\begin{problem}
[{\sc Machine Reading at Scale; MRS}]
\label{prob:qa}
Given a question, an MRS system retrieves $k$ passages relevant to the question in a corpus $D$ (IR task) and extracts an answer from the retrieved passages (RC task).
\end{problem}

\begin{definition}
A \textit{question}, $q$, is a sentence in natural language.
\end{definition}

\begin{definition}
A \textit{passage}, $x$, is a short part of a document in natural language. It does not contain any non-textual information. 
\end{definition}

\begin{definition}
A \textit{corpus}, $D$, is a collection of passages.
\end{definition}

\begin{definition}
An \textit{answer} is a span of arbitrary length within a passage. Its type is not limited to single words or named entities. It is extracted (not synthesized and generated) from the passage.
\end{definition}

\begin{definition}
\label{def:relevant}
A \textit{relevant passage} to a question is one that contains all textual facts required to answer the question. The IR task requires such relevant passages to be found.
\end{definition}

\section{Proposed model}
Our \textit{Retrieve-and-Read} model is based on the bi-directional attention flow (BiDAF) model~\cite{SeoKFH17}, which is a standard RC model. As shown in Figure~\ref{fig:network}, it consists of six layers:

\begin{figure}[t!]
\centering
\includegraphics[width=.45\textwidth]{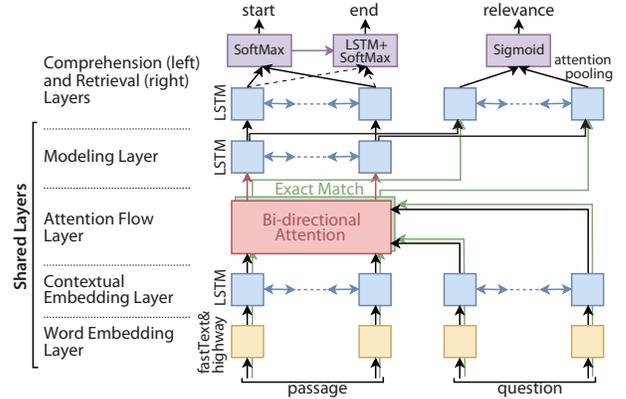}
\caption{Retrieve-and-Read model architecture.}
\label{fig:network}
\end{figure}

\begin{layer}
The \textbf{word embedding layer} maps all words to a vector space using pre-trained word embeddings. 
\end{layer}
\begin{layer}
The \textbf{contextual embedding layer} encodes the temporal interactions between words.
\end{layer}
\begin{layer}
The \textbf{attention flow layer} couples question and passage vectors and produces a sequence of question-aware passage word vectors.
\end{layer}
\begin{layer}
The \textbf{modeling layer} captures the interaction among passage words conditioned on the question.
\end{layer}
\begin{layer}
The \textbf{comprehension layer} outputs an answer to the question.
\end{layer}
\begin{layer}
The \textbf{retrieval layer} provides the relevance of the passage. 
\end{layer}

\noindent The first four layers of our model are shared by the IR and RC tasks, and it has a new task-specific layer for the IR task. The model jointly learns the two tasks by combining their loss functions. In addition to the attention mechanism in the shared layers, the retrieval layer calculates a binary exact-match channel to capture the question intent clearly and has a self-attention mechanism to retain important word representations for retrieval. We note that the RC component trained with single-task learning (STL) is essentially equivalent to BiDAF, except for the word embedding layer that has been modified to improve accuracy.

\subsection{Word embedding layer}
Let $x = \{x_1, \ldots, x_T\}$ and $q = \{q_1, \ldots, q_J\}$ represent one-hot vectors of words in the input passage and question. This layer projects each of the one-hot vectors (size of $V$) into a $v$-dimensional continuous vector space with a weight matrix $W^e \in \mathbb{R}^{v \times V}$.

The embedding vectors are passed to a two-layer highway network~\cite{SrivastavaGS15} that is shared for the question and passage. The outputs are two sequences of $v$-dimensional vectors: $X \in \mathbb{R}^{v \times T}$ for the passage ($T$ words) and $Q \in \mathbb{R}^{v{\times}J}$ for the question ($J$ words).

Note that the original BiDAF uses a pre-trained GloVe~\cite{PenningtonSM14} and also trains character-level embeddings by using a CNN~\cite{Kim14} in order to handle out-of-vocabulary (OOV) or rare words. Instead of using GloVe and CNN, our model uses fastText~\cite{BojanowskiGJM17} for the fixed pre-trained word vectors and removes character-level embeddings. The fastText model takes into account subword information and can obtain valid representations even for OOV words.

\subsection{Contextual embedding layer}
This layer has a single-layer LSTM~\cite{HochreiterS97} in the forward and backward directions and concatenates the $d$-dimensional outputs of the two LSTMs. It has learnable parameters for both directions. It obtains $H \in \mathbb{R}^{2d{\times}T}$ from $X$ and $U \in \mathbb{R}^{2d{\times}J}$ from $Q$.

\subsection{Attention flow layer}
This layer computes attentions in two directions in order to fuse information from the passage (i.e., context) to the question (i.e., query) words as well as from the question to the passage. It first computes a similarity matrix $S \in \mathbb{R}^{T{\times}J}$, 
\begin{align}
S_{tj} = {w^s}^{\mathsf{T}} [ H_t; U_j; H_t \odot U_j ],
\end{align}
that indicates the similarity between the $t$-th passage word and the $j$-th question word. $w^s \in \mathbb{R}^{6d}$ are learnable parameters, the $\odot$ operator denotes the Hadamard product, and the $[; ]$ operator is vector concatenation across the rows.

\textbf{Context-to-query attention} signifies which question words are most relevant to each passage word. It outputs $\tilde{U}_t = \sum_j a_{tj} U_j \in \mathbb{R}^{2d}$, where $a_t = \mathrm{softmax}_j(S_{t}) \in \mathbb{R}^J$.

\textbf{Query-to-context attention} signifies which passage words have the closest similarity to one of the question words. It outputs $\tilde{h} = \sum_t b_t H_t \in \mathbb{R}^{2d}$, where $b = \mathrm{softmax}_t(\mathrm{max}_j(S)) \in \mathbb{R}^T$. It then obtains $\tilde{H} \in \mathbb{R}^{2d{\times}T}$ by tiling the vector $T$ times across the columns.

\textbf{Bi-directional attention} computes $G$ to obtain a question-aware representation of each passage word,
\begin{align}
G = [ H; \tilde{U}; H \odot \tilde{U}; H \odot \tilde{H} ] \in \mathbb{R}^{8d{\times}T}.
\end{align}

\subsection{Modeling layer}
This layer uses a single-layer bi-LSTM and obtains $M \in \mathbb{R}^{2d{\times}T}$ from $G$. The output is passed on to the task-specific layers.

\subsection{Comprehension layer}
The RC task requires the model to find a phrase to answer the question. 
This layer uses the concept of pointer networks \cite{VinyalsFJ15}, 
where the phrase is derived by predicting the start and end indices in the passage.

First, this layer passes the output of the modeling layer $M$ to another single-layer bi-LSTM and obtains $M^{1} \in \mathbb{R}^{2d{\times}T}$. It then calculates the probability distribution of the start index,
\begin{align}
\label{eq:p1}
p^1 = \mathrm{softmax}_t( {w^1}^{\mathsf{T}} [G; M^{1}]) \ \in \mathbb{R}^{T},
\end{align}
where $w^1 \in \mathbb{R}^{10d}$ are learnable parameters.

Next, it calculates an attention-pooling vector, $\tilde{m}^{1} = \sum_t p^1_t M^{1}_t \in \mathbb{R}^{2d}$, and tiles it $T$ times to obtain $\tilde{M}^1 \in \mathbb{R}^{2d{\times}T}$. Then, it passes the concatenated matrix of $[G; M^{1}; \tilde{M}^{1}; M^{1} \odot \tilde{M}^{1}] \in \mathbb{R}^{14d{\times}T}$ to another single-layer bi-LSTM and obtains $M^{2} \in \mathbb{R}^{2d{\times}T}$.

Finally, it calculates the probability distribution of the end index,
\begin{align}
\label{eq:p2}
p^2 = \mathrm{softmax}_t({w^2}^\mathsf{T} [G; M^{2}])  \in \mathbb{R}^{T},
\end{align}
where $w^2 \in \mathbb{R}^{10d}$ are learnable parameters.

\subsection{Retrieval layer}
The IR task requires the model to find relevant passages that meet the information needs of the question. This layer maps the output of the modeling layer to the relevance score of the passage 
by using a binary exact-matching and a self-attention mechanism.

First, for capturing question words clearly, this layer calculates a binary exact-match channel, $\tilde{B}$. Let $B \in \mathbb{R}^{J{\times}T}$ be a matrix such that the entry at position $(j, t)$ is 1 if the $j$-th question word is an exact match to the $t$-th passage word in the passage and 0 otherwise. It performs a max-pooling of $B$ over all the question words to obtain ~$\tilde{B} \in \mathbb{R}^{1{\times}T}$.

Then, it passes $[M; \tilde{B}] \in \mathbb{R}^{(2d+1){\times}T}$ to another single-layer bi-LSTM and obtains $M^{r} \in \mathbb{R}^{2d{\times}T}$. To retain important word representations for retrieval, it calculates an attention-pooling of $M^{r}$,
\begin{align}
\tilde{m}^{r} = \sum\nolimits_t \beta_t M^{r}_t  \ \in \mathbb{R}^{2d}.
\end{align}
The element of the attention $\beta \in \mathbb{R}^T$ is computed as the inner product between a question-aware representation of each passage word and a self-attention context vector, $w^c$: 
\begin{align}
\beta_t = \exp({m_t}^\mathsf{T} w^c) / \sum\nolimits_{t'}\exp({m_{t'}}^\mathsf{T} w^c),
\end{align}
where $m_t = W^{a} M^r_t + b^{a}$. $W^a \in \mathbb{R}^{c{\times}2d}$ and $b^a, w^c \in \mathbb{R}^{c}$ are learnable parameters.

Finally, it calculates the relevance of the passage to the question,
\begin{align}
p^r = \mathrm{sigmoid}({w^r}^\mathsf{T} \tilde{m}^r) \ \in [0,1],
\end{align}
where $w^r \in \mathbb{R}^{2d}$ are learnable parameters.

\subsection{Multi-task learning}
We define the training loss as the sum of losses in IR and RC, 
\begin{align}
L(\theta) =  L_{RC} + \lambda L_{IR}, 
\end{align}
where $\theta$ is the set of all learnable parameters of the model and $\lambda$ is a balancing parameter. Our model uses question-answer-passage triples in the training set as positive examples and generates negative examples from the set. We explain the negative sampling procedure used in our experiments in Section~\ref{sec:negative}.

The loss of the IR task, $L_{IR}$, is the binary cross entropy between the true and predicted relevance scores averaged over all examples:
\begin{align}
L_{IR} = -\frac{1}{N} \sum\nolimits_i \left(r_i \log p^r + (1 - r_i) \log (1 - p^r) \right),
\end{align}
where $N$ is the number of examples and $r_i$ is the true relevance score (1 if the $i$-th example is positive, 0 otherwise).

The loss of the RC task, $L_{RC}$, is the negative log probabilities of the true start and end indices given by the predicted distributions averaged over all \textit{positive} examples:
\begin{align}
L_{RC} = - \frac{1}{N_{\rm pos}} \sum\nolimits_i  r_i \left( \log p^1_{y_i^1} + \log p^2_{y_i^2} \right),
\end{align}
where $N_{\rm pos}$ is the number of positive examples, and $y_i^1$ and $y_i^2$ are the true start and end indices. Note that negative examples are ignored in the loss function for RC because they do not have the correct answer spans for the query.

\subsection{Test process}
\label{sec:test-process}
\subsubsection{IR task.}
Given a question $q$, our model outputs the top-$k$ passages, $R_k$, ordered by the relevance $p^r$ for each passage $x \in D$ and $q$.

\subsubsection{RC task.}
Our model outputs an answer for each passage in the retrieved set. In total, it outputs $k$ answers. Given $x \in R_k$ and $q$, it chooses the answer span $(t_1, t_2)$ where $t_1 \leq t_2$ with the maximum value of $p^1_{t_1} p^2_{t_2}$, which can be computed in linear time with dynamic programming. It then outputs an answer as a substring of $x$ with the chosen span. 

\subsubsection{MRS task.}
\label{sec:voting}
Our model returns a final answer with weighted majority voting from the $k$ outputs of the RC component. It uses $\exp(p^r / \tau)$ for each RC output as a weight, where $\tau$ is a temperature parameter that controls the voting. It sums the weights of each answer string and selects the most voted for answer as the final answer. Note that we do not use the RC score for the voting.

\subsection{Telescoping setting}
\label{sec:telescoping}
It is difficult to adapt neural networks to the whole of a large-scale corpus due to their computational cost; so, we consider using a \textit{telescoping} setting that uses chaining of different IR models, where each successive model re-ranks a smaller number of passages~\cite{MatveevaBBLW06}. 

Without loss of generality, we can use a telescoping setting with our model, where our IR component finds relevant passages in a subset of a corpus $D$ retrieved by chaining of different IR models. That is, the initial rankers focus on eliminating higher irrelevant passages, and our model operates as a re-ranker for determining the existence of answer phrases within the remaining passages. We explain the settings used in our experiments in Section~\ref{sec:teles}. 

\section{Experiments}
\subsection{Datasets}
\subsubsection{Training} 
We used \textbf{SQuAD} 1.1~\cite{RajpurkarZLL16}, which is a standard RC dataset based on Wikipedia articles, and \textbf{Jp-News}, which was created by crowdworkers in the same way as SQuAD. For each question, we defined that the single passage corresponding to the question in the dataset is relevant and the others are not relevant. That is, our IR component used question-answer-passage triples in the training set as positive examples. It generated negative examples from the set, as described in Section~\ref{sec:negative}.

Table~\ref{tb:dataset} shows the statistics of the datasets. The main characteristic of Jp-News is the use of a set of Japanese news articles that contain similar passages on the same topics. These similar passages make it difficult for IR models to find the most relevant passage to each question.
See the Appendix~\ref{sec:dataset} for details of the data collection and analysis of the Jp-News dataset.

\subsubsection{Evaluations}
We used a benchmark for MRS, \textbf{SQuAD}$_\mathrm{full}$~\cite{ChenFWB17}, to evaluate our model trained with SQuAD. It takes only the question-answer pairs of the SQuAD development set and uses the full Wikipedia as the knowledge source (5,075,182 articles). In this \textit{end-to-end} setting, no relevance information for IR is given. To evaluate the \textit{individual} contributions of our model components precisely, we conducted additional experiments using the passages of the SQuAD development set (denoted as \textbf{SQuAD}$_\mathrm{dev}$) or those of Jp-News as the knowledge source. 

As in the training set, the question-answer-passage triples in the development set were used as positive examples. Our IR evaluation using binary relevance scores is more rigorous than distant-supervision evaluations based on whether the ground-truth appears in the retrieved passages~\cite{ChenFWB17,WangAAAI2018}.

\begin{table}[t!]
\centering
\caption{Number and mean length (in tokens) of each item in the training datasets.}
\label{tb:dataset}
\begin{tabular}{l|cc|ccc}\hline
& \multicolumn{2}{c|}{SQuAD} & \multicolumn{3}{c}{Jp-News} \\ 
& train & dev & train & dev & test \\ \hline
No. articles & 442 & 48 & 4,000 & 500 & 500 \\
No. questions & 87,599 & 10,570 & 66,073 & 8,247 & 8,272 \\
No. passages & 18,896 & 2,067 & 10,024 & 1,214 & 1,247 \\ 
No. answers & 87,599 & 34,726 & 179,908 & 22,500 & 22,500 \\ 
Len. questions & 11.4 & 11.5 & 21.9 & 21.8 & 21.9 \\
Len. passages & 140.3 & 144.5 & 181.4 & 176.2 & 177.7 \\ 
Len. answers & 3.5 & 3.3 & 4.3 & 4.5 & 4.2 \\ 
\hline
\end{tabular}
\end{table}

\subsection{Evaluation metrics}
\subsubsection{IR Task}
We used metrics for binary relevance judgments to evaluate the individual contributions of our IR component.
\textbf{Success@k} (S@k) is the percentage of times that a relevant document is in the top-$k$ retrieved documents, $R_k$, for a query~\cite{Craswell09e}.
\textbf{MRR@k} (M@k) is the mean reciprocal rank of the first relevant document~\cite{Craswell09a}.

\subsubsection{RC and MRS Tasks}
We evaluated the models with the same metrics used in SQuAD.
\textbf{EM} (Exact match) measures the percentage of predictions that match the ground truth exactly.
\textbf{F1} (Macro-averaged F1 score) measures the average overlap between the bag of words of the prediction and that of the ground truth~\cite{RajpurkarZLL16}.

\subsection{Baselines}
\subsubsection{MRS task}
We used two state-of-the-art models: \textbf{DrQA}~\cite{ChenFWB17} and \textbf{R$^3$}~\cite{WangAAAI2018} in the end-to-end setting. Moreover, for the detailed evaluations, we used a simple pipeline of a TF-IDF model (without re-ranking) and our RC component trained with STL. This structure corresponds to the one of DrQA~\cite{ChenFWB17}. We also evaluated a pipeline of our IR component trained with STL (used as a re-ranker) and our RC component trained with STL.

\subsubsection{IR task}
For the individual evaluations, we used two recent neural IR models as re-ranker baselines. \textbf{Duet} \cite{MitraDC17} is a recent standard neural IR model. It consists of two separate CNNs: one that exactly matches question and passage words and another that matches the question and the passage by using learned distributed representations. In \cite{MitraDC17}, \citeauthor{MitraDC17} reported that Duet significantly outperformed non-neural IR models such as BM25~\cite{RobertsonW94} and LSA~\cite{DeerwesterDLFH90} and also earlier neural IR models such as DSSM~\cite{HuangHGDAH13} and DRMM~\cite{GuoFAC16}. \textbf{Match-tensor} \cite{JaechKRC17} is a recent model that uses RNNs for the encoding of the input query and passage. It uses soft-matching between each question and passage word encoded by a bi-LSTM and uses 2D convolutions that map the matching tensor to the relevance score. Although Duet does not assume that queries are given in the form of natural language, Match-tensor can carefully consider the context of question and passage sentences with RNNs.

\subsubsection{RC task}
In order to confirm that our MTL approach does not degrade RC performance, we used \textbf{BiDAF}~\cite{SeoKFH17}, which is a base model of our model, and \textbf{Document Reader}~\cite{ChenFWB17}, which is the RC component of DrQA.

\subsection{Model configuration}
\label{sec:config}
\subsubsection{Preprocess}
We used the Stanford CoreNLP tokenizer \cite{ManningSBFBM14} (JTAG tokenizer \cite{FuchiT98}) on the SQuAD (Jp-News) dataset. Our model used pre-trained $300$-dimensional fastText embeddings~\cite{BojanowskiGJM17} in a case-sensitive manner, and they were fixed during training. We used the 2016-03-05 (2017-11-03) dump of English (Japanese) Wikipedia articles for pre-training.

\subsubsection{Training process}
We used the same configuration for all datasets. We trained our model with 10 GPUs (GeForce GTX 1080 Ti). Each GPU processed a minibatch of size 60, consisting of 30 positive and 30 negative examples. The LSTM hidden size, $d$, and the context vector size, $c$, were set to 100. Weights were initialized using the Xavier uniform initializer~\cite{GlorotB10}, except that the biases of all the linear transformations were initialized with zero vectors. A dropout \cite{SrivastavaHKSS14} rate of 0.2 was used for all highway and LSTM layers and each linear transformation before the softmax and sigmoid for the outputs. We used SGD with a momentum of 0.9 and an initial learning rate of 1. The number of epochs was 15, and the learning rate was reduced by a factor of 0.9 every epoch. 
The balancing factor of MTL, $\lambda$, was set to 1.
During training, the moving average of each weight was maintained with an exponential decay rate of 0.99. At test time, the moving average was used instead of the raw weight. Single-task learning (STL) was conducted by changing the training loss function. We used $L(\theta)=L_{IR}$ for the IR task and $L(\theta)=L_{RC}$ for the RC task.

\subsubsection{Negative sampling}
\label{sec:negative}
Negative examples for training were generated from positive examples. Each negative example consisted of the same question and a similar passage, which was randomly sampled among the top-15 most similar passages in a TF-IDF vector space of the training set, to the corresponding positive example. Preliminary experiments showed that negative examples consisting of a question and a passage that were randomly sampled from the whole training set were not effective at training IR modules.

\subsubsection{Telescoping settings}
\label{sec:teles}
We used two settings: T1 and T2. 

\textbf{T1}.\, For SQuAD$_\mathrm{full}$, we used chaining of two exact-matching IR models and one neural IR model. The first model retrieved five articles from about five million articles, and the second one retrieved $200$ passages from the five articles. Articles were split into passages by one or more line breaks, as in \cite{ChenFWB17}. We used Document Retriever~\cite{ChenFWB17}, which is a model based on bigram hashing and TF-IDF matching, for both the first and second retrievals. Finally, the IR component of our model found the top-$1$ passage from the 200 passages and passed it to our RC component. 

\textbf{T2}.\, For the individual evaluations using SQuAD$_\mathrm{dev}$ and Jp-News, one TF-IDF model retrieved $200$ passages from the whole passages in the evaluation set, and our neural IR component retrieved the top-$k$ passage. $k$ was varied from $1$ to $5$.
The temperature parameter for voting, $\tau$, was set to $0.05$.

\subsubsection{Baseline settings.}
For MRS and RC baselines with the SQuAD dataset, we used the results reported in their studies.
For Jp-News, we trained and evaluated BiDAF and Document Reader using the original configuration of each study. We trained GloVe embeddings with the same Wikipedia articles that our model used for pre-training. We did not use the lemma, POS, or NER features for Document Reader, because they degraded accuracy. 

The IR baselines used the same telescoping settings as our model. We used the original configuration of each IR method, except as follows. We used the $300$-dimensional fastText (which our model used) for the fixed embeddings. Although the original Duet uses character $n$-grams for learning the word embeddings, it does not work well when there is not much training data. 

\subsection{Results}
The reported results of all neural models with different initializations are means over five trials.

\subsection*{\normalsize Does our system achieve state-of-the-art performance on a large-scale MRS task?}

\begin{table}[t!]
\centering
\tabcolsep=3pt
\caption{MRS using full Wikipedia results. S, DS, DS', and E mean supervised learning, distant supervision with SQuAD and with other datasets, and ensemble model, respectively. The results of our single model and R$^3$ are averages of five runs; the superscript is the standard error.} 
\label{tb:fullMRS}
\begin{tabular}{lcccc|cc}\hline
& & & & & \multicolumn{2}{c}{SQuAD$_\mathrm{full}$} \\
& S & DS & DS' & E & EM & F1 \\ \hline
DrQA & \checkmark& & & & 27.1 & -- \\
DrQA+DS & \checkmark& \checkmark & & & 28.4 & -- \\
DrQA+DS+MTL & \checkmark& \checkmark & \checkmark & & 29.8 & -- \\
$\mathrm{R}^3$ & & \checkmark & & &29.1$^{.2}$ & 37.5$^{.2}$ \\ \hline
Retrieve-and-Read (single) & \checkmark& & & & \textbf{32.7}$^{.2}$ & \textbf{39.8}$^{.2}$ \\
Retrieve-and-Read (ensemble) & \checkmark& & & \checkmark & \textbf{35.6} & \textbf{42.6} \\\hline
\end{tabular}
\end{table}

We evaluated the overall performance of our single and ensemble models with SQuAD$_\mathrm{full}$ and the T1 telescoping setting. 
The ensemble model consists of five training runs with the identical architecture and hyper-parameters.
It chooses the answer with the highest sum of confidence scores amongst the five runs for each question.
Table~\ref{tb:fullMRS} shows that our models outperformed the state-of-the-art by a significant margin. 
The improvement of our single (ensemble) model
over R$^3$, which was trained without using answer spans, 
was up to 3.6\% (6.5\%) in EM and 2.3\% (5.3\%) in F1. 
This result indicates the effectiveness of learning from answer spans in IR.

\subsection*{\normalsize Does our MTL improve the accuracy of STL, which does not consider answer spans, in IR?}
The individual contributions of our neural IR component on SQuAD$_\mathrm{dev}$ and Jp-News were evaluated using 
the T2 telescoping setting. Table~\ref{tb:IRresult} shows that our IR component trained with MTL significantly outperformed STL. The IR component shares hidden layers with the RC component in order that it can learn from answer spans, and this sharing contributed to statistically significant improvements over all baselines ($t$-test; $p<.001$) for all datasets. 
\begin{table}[t!]
\centering
\tabcolsep=4.5pt
\caption{Averaged IR (re-ranking) results. The initial ranker was TF-IDF.}
\label{tb:IRresult}
\begin{tabular}{l|cc|cccc}\hline
& \multicolumn{2}{c}{SQuAD$_\mathrm{dev}$} & \multicolumn{2}{|c}{Jp-News$_\mathrm{dev}$} & \multicolumn{2}{c}{Jp-News$_\mathrm{test}$} \\

IR re-ranker & S@1 & M@5 & S@1 & M@5 & S@1 & M@5\\ \hline
(None) & 0.748 & 0.810 & 0.713 & 0.824 & 0.692 & 0.804 \\ 
Duet & 0.665 & 0.743 & 0.573 & 0.698 & 0.564 & 0.692 \\
Match-tensor & 0.732 & 0.791 & 0.725 & 0.821 & 0.704 & 0.806 \\ \hline
Our IR (STL) & 0.707 & 0.773 & 0.690 & 0.800 & 0.673 & 0.787 \\
Our IR (MTL) & {\bf 0.811} & {\bf 0.863} & {\bf 0.753} & {\bf 0.842} & {\bf 0.737} & {\bf 0.830 } \\
\hline
\end{tabular}
\end{table}

Other re-rankers did not clearly outperform TF-IDF. Interestingly, our IR component trained with STL performed significantly worse than TF-IDF and Match-tensor. This result indicates that it is important to teach which part of the passage meets the information needs rather than to give only a relevance score to the whole passage  and that our MTL approach allows for accurate learning from a small amount of data. We should note that the experiments conducted on Duet and Match-tensor in their original studies used a set of approximately one million documents, so they would outperform TF-IDF when there is a large amount of data.

\subsection*{\normalsize Does our MTL improve the accuracy of STL in RC?}

\begin{table}[t!]
\centering
\caption{Averaged standard RC (reading one relevant passage) results for a single model.}
\label{tb:RCresult}
\begin{tabular}{l|cc|cccc}\hline
& \multicolumn{2}{c}{SQuAD$_\mathrm{dev}$} & \multicolumn{2}{|c}{Jp-News$_\mathrm{dev}$} & \multicolumn{2}{c}{Jp-News$_\mathrm{test}$} \\
RC model & EM & F1 & EM & F1 & EM & F1 \\ \hline
BiDAF & 67.7 & 77.3 & 76.9 & 88.1 & 77.3 & 88.3\\
Document Reader & {\bf 69.5} & {\bf 78.8} & 75.9 & 87.6 & 76.2 & 87.8\\ 
\hline 
Our RC (STL) & 69.1 & 78.2 & 77.4 & 88.4 & 78.3 & 88.8\\
Our RC (MTL) & 69.3 & 78.5 & {\bf 78.0} & {\bf 88.8} & {\bf 78.8} & {\bf 89.2}\\ \hline
\end{tabular}
\end{table}

\begin{figure}[t!]
\centering
\includegraphics[width=.47\textwidth]{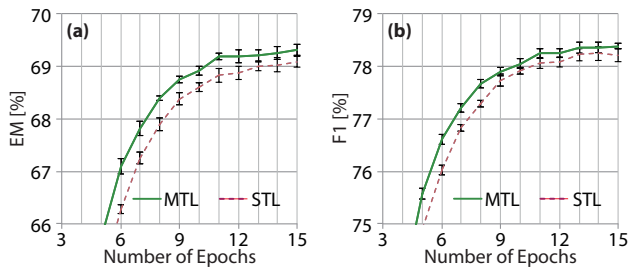}
\caption{Averaged learning curves of our RC component trained with MTL or STL. (a) EM and (b) F1 metrics on SQuAD$_\mathrm{dev}$ set. Error bars are for over five trials.}
\label{fig:loss_RC}
\end{figure}

We evaluated the individual contributions of our RC component using with SQuAD$_\mathrm{dev}$ and Jp-News. Table~\ref{tb:RCresult} and Figure~\ref{fig:loss_RC} show the results for the \textit{standard} RC task, where each model was given one relevant passage for each question. As shown in Figure~\ref{fig:loss_RC}, our MTL approach performed statistically significantly better than STL in terms of EM and F1 of each epoch (Two-way repeated-measures ANOVA; $p<.05$) for all datasets. Although our RC component based on a vanilla BiDAF was not competitive among the current state-of-the-art methods such as QANet~\cite{Yu18}, we confirmed that our MTL does not degrade RC performance and it is comparable to the Document Reader model, which is used in DrQA~\cite{ChenFWB17}.

\subsection*{\normalsize Does our MTL improve the accuracy of pipeline approaches in MRS?}
We compared our MTL approach with a pipeline of our components trained with STL, by using SQuAD$_\mathrm{dev}$ and Jp-News. We also evaluated a simple pipeline of TF-IDF finding the top-1 passage and our RC component trained with STL. Table~\ref{tb:QAresult} shows that our MTL approach with the T2 telescoping setting ($k=1$) statistically significantly outperformed the pipeline approaches ($t$-test; $p<.001$) for all datasets. As shown in Tables \ref{tb:IRresult} and \ref{tb:RCresult}, the improvements in our IR component were responsible for this progress. 

\begin{table}[t!]
\centering
\caption{MRS results of our single and ensemble models. The initial ranker was TF-IDF. (IR, RC) = (None, STL) corresponds to the pipeline of TF-IDF and a neural RC model~\cite{ChenFWB17}.}
\label{tb:QAresult}
\begin{tabular}{cc|cc|cccc}\hline
\multicolumn{8}{l}{Single model~(mean performance over five trials)} \\ \hline
& & \multicolumn{2}{c}{SQuAD$_\mathrm{dev}$} & \multicolumn{2}{|c}{Jp-News$_\mathrm{dev}$} & \multicolumn{2}{c}{Jp-News$_\mathrm{test}$} \\

Our IR & Our RC & EM & F1 & EM & F1 & EM & F1 \\ \hline
(None) & STL & 53.9 & 61.6 & 65.6 & 78.0 & 65.6 & 77.9 \\
STL & STL & 52.2 & 59.9 & 64.9 & 77.2 & 65.5 & 77.7 \\
MTL & MTL & {\bf 60.0} & {\bf 68.1} & {\bf 69.5} & {\bf 81.7} & {\bf 70.6} & {\bf 82.7} \\ \hline \hline
\multicolumn{8}{l}{Ensemble model consisting of five single models} \\ \hline
Our IR & Our RC & EM & F1 & EM & F1 & EM & F1 \\ \hline
(None) & STL & 56.6 & 63.6 & 68.2 & 79.7 & 67.9 & 79.4 \\
STL & STL & 56.3 & 63.2 & 68.8 & 80.2 & 68.8 & 80.4 \\
MTL & MTL & {\bf 64.5} & {\bf 71.8} & {\bf 73.5} & {\bf 84.5} & {\bf 75.0} & {\bf 85.9} \\ \hline
\end{tabular}
\end{table}

\subsection*{\normalsize Does our whole system run at a practical speed?}
The test process for SQuAD$_\mathrm{dev}$ using our single model with (without) the T2 telescoping setting, where our neural model processed 200 (2,047) passages for each question, took 1.5 (17.1) seconds per question. The time taken by the TF-IDF search was very short (less than 10 milliseconds). Also, the test process for SQuAD$_\mathrm{full}$, in which the mean length of the passages is shorter than in that of SQuAD$_{\mathrm{dev}}$, took 1.0 second per question when using our single model with the T1 telescoping setting.

To summarize, the whole system could run at a practical speed under the telescoping settings, and the computational order of the system was dependent on the size of the subset retrieved by TF-IDF and on the lengths of the question and passages. 

\subsection*{\normalsize Does the telescoping setting degrade accuracy in IR?}
We evaluated the accuracy of the initial ranker, TF-IDF, on the SQuAD$_\mathrm{dev}$ and Jp-News datasets, when it retrieved 200 passages in the T2 telescoping setting. We confirmed that it eliminated higher irrelevant passages with almost no deterioration in accuracy: the Success@200 rate was 0.991 (0.997) on the SQuAD (Jp-News) development set, while the Success@1 rate was 0.748 (0.713). 

Although TF-IDF was reasonable as an initial ranker on these datasets, we need to improve the accuracy to overcome the problem of lexical variation between the question and the passages. We will discuss this issue in Section~\ref{sec:related}. 

\subsection*{\normalsize Detailed analysis}
\begin{figure}[t!]
\centering
\includegraphics[width=.44\textwidth]{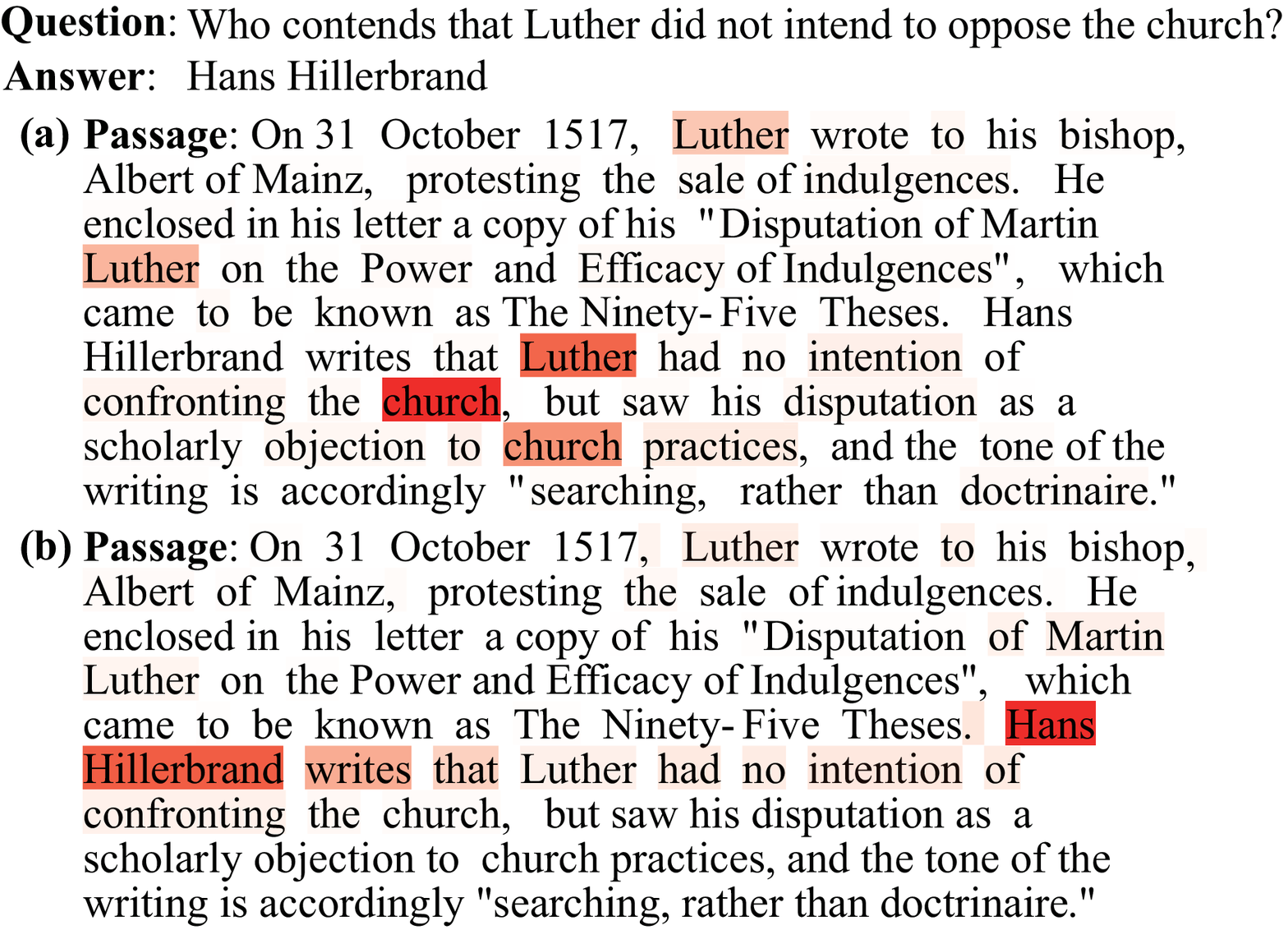}
\caption{Attention of our IR components trained with (a) STL and (b) MTL to passage words. Darker red signifies greater attention.}
\label{fig:vis2}
\end{figure}

Figure~\ref{fig:vis2} provides a qualitative analysis of our IR components' attention to a passage. The component trained with MTL captured the answer phrase as well as question words, but the component trained with STL did not recognize the answer phrase.
{}

\begin{figure}[t!]
\centering
\includegraphics[width=.44\textwidth]{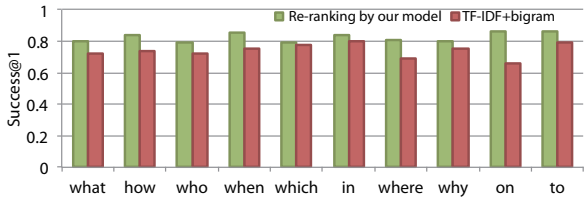}
\caption{Success@1 rates broken down by the ten most-frequent first words in the question on the SQuAD dev.~set.}
\label{fig:qtype_ir}
\end{figure}

Figure~\ref{fig:qtype_ir} shows IR results broken down by the first words in the question on SQuAD$_\mathrm{dev}$. Our IR component trained with MTL performed better than TF-IDF in every category.

Table~\ref{tb:ablation} shows the accuracy of our IR component with and without the exact-match channel on SQuAD$_\mathrm{dev}$ and Jp-News. This channel improved the model performance ($t$-test; $p<.01$).
\begin{table}[t!]
\centering
\tabcolsep=2.5pt
\caption{Ablation test of our IR re-ranker model. The results are averages over five trials.}
\label{tb:ablation}
\begin{tabular}{l|cc|cccc}\hline
& \multicolumn{2}{c}{SQuAD$_\mathrm{dev}$} & \multicolumn{2}{|c}{Jp-News$_\mathrm{dev}$} & \multicolumn{2}{c}{Jp-News$_\mathrm{test}$} \\
IR re-ranker & S@1 & M@5 & S@1 & M@5 & S@1 & M@5\\ \hline
Our IR (MTL) & {\bf 0.811} & {\bf 0.863} & {\bf 0.753} & {\bf 0.842} & {\bf 0.737} & {\bf 0.830 } \\
w/o exact-match & 0.800 & 0.858 & 0.742 & 0.832 & 0.726 & 0.820 \\ 
\hline
\end{tabular}
\end{table}

Table~\ref{tb:voting} shows the results of our model using the T2 telescoping setting with different values of $k$, which is the number of passages retrieved by our IR component. The results for $k$ = 1 were the best for corpora like SQuAD (Wikipedia articles), where most descriptions are expected to be stated only once, while larger values of $k$ were suitable for corpora like Jp-News that include the same descriptions in multiple passages.
\begin{table}[t!]
\centering
\tabcolsep=4pt
\caption{Effect of varying the number of retrieved passages, $k$, on the averaged MRS results of our single model.}
\label{tb:voting}
\begin{tabular}{l|cccc|cccc}\hline
& \multicolumn{2}{c}{SQuAD$_\mathrm{full}$} & \multicolumn{2}{c}{SQuAD$_\mathrm{dev}$} & \multicolumn{2}{|c}{Jp-News$_\mathrm{dev}$} & \multicolumn{2}{c}{Jp-News$_\mathrm{test}$} \\
$k$ & EM & F1 & EM & F1 & EM & F1 & EM & F1 \\ \hline
1 &{\bf 32.7} & {\bf 39.8} & {\bf 60.0} & {\bf 68.1} & 69.5 & 81.7 & 70.6 & 82.7\\ 
3 & 32.4 & 39.5 & 59.5 & 67.6 & 69.7 & 81.9 & 70.8 & 82.8\\ 
5 & 31.4 & 38.4 & 58.9 & 67.0 & {\bf 69.9} & {\bf 82.0} & {\bf 70.9} & {\bf 82.9}\\ \hline
\end{tabular}
\end{table}

The total number of parameters without word embeddings in our MTL (STL = IR + RC) model was 3.11M (4.52M = 1.67M + 2.85M). The MTL model shared hidden layers for the two tasks and could save 1.41M parameters compared with the pipeline system of the two individual components. Also, the training time of MTL with SQuAD was 14.0 hours. It could save 4.6 hours compared with the pipeline system (8.6 hours for IR and 10.0 hours for RC components). 

\section{Related work and discussion}
\label{sec:related}
\paragraph{Machine reading at scale} MRS, which is a combination of an IR and RC task that uses a large number of unstructured texts as a knowledge source, is an emerging research field. As described earlier, the work of \citeauthor{WangAAAI2018}~\cite{WangAAAI2018} is the most similar to ours. In their model, R$^3$, the ranker and reader share the hidden layers and they are jointly learned with reinforcement learning. The largest difference between DrQA~\cite{ChenFWB17} and our model is that R$^3$ was trained entirely using distant supervision. Although distant supervision without using answer \textit{spans} can learn from a large amount of data, it may prevent models from carefully considering the context of passages. We believe that supervised learning using answer spans is also promising because of its high accuracy. Our model can be trained with standard RC datasets with answer spans, and such datasets can be collected by crowdsourcing. 

Recently, \citeauthor{WangICLR2018} proposed answer re-ranking methods that reorder the answer candidates generated by the RC module of an MRS system~\cite{WangICLR2018}. Their methods can be used as post-processes for existing MRS systems including ours; unfortunately, they did not report their performance on SQuAD$_\mathrm{full}$. Their methods (and our weighted voting scheme described in Section~\ref{sec:voting}) are effective when the correct answer is repeatedly suggested in passages. However, there are still problems with 
question answering that combines disjoint pieces of textual evidence. 
Effective methods of text understanding across multiple passages need to be developed that would alleviate the limitations of our Definition~\ref{def:relevant}, which assumes that all textual facts required to answer the question are contained in one relevant passage.

Currently, SQuAD$_{\mathrm{full}}$ is the only large-scale MRS dataset that can be used to both train and evaluate \textit{extractive} RC models in the \textit{same domains and question styles}. 
NewsQA~\cite{TrischlerWYHSBS16} can be used as the training data for extractive RC models, while its questions are too dependent on the corresponding passage (e.g., Who is hiring?)~to use the MRS task.
\citeauthor{Clark18} recently released the ARC dataset as a more challenging dataset~\cite{Clark18}. It consists of 7,787 QA pairs and 14M science sentences, although its multiple-choice type of question is different from that of SQuAD and JP-News.

\paragraph{RC with a small number of passages} Several small-scale MRS datasets consisting of sets of (question, answer string, evidence passages) triples have been proposed: TriviaQA~\cite{JoshiCWZ17}, Quasar-T~\cite{DhingraMC17}, SearchQA~\cite{DunnSHGCC17} (\textit{answer extraction}), WikiHop~\cite{WelblSR17} (\textit{multiple choice}), Quasar-S~\cite{DhingraMC17} (\textit{cloze-styles}), and MS~MARCO~\cite{NguyenRSGTMD16} (\textit{answer generation}). Each QA pair in these datasets has a \textit{small} number of evidence passages that were gathered automatically from the Web by using a search engine as their initial ranker in a large-scale MRS setting. 

Approaches that read all given passages at once are often adapted to work with these datasets. That is, IR methods are not used. For example, R-NET for MS MARCO~\cite{WangYWCZ17} concatenates all ten passages corresponding to a question and extracted an answer span from the concatenated one. However, such approaches consume a lot of memory and do not work well when there are many long passages.

Moreover, the datasets listed above do not provide answer spans in evidence passages and cannot be used for \textit{supervised training} of our model because of the lack of a relevance score and answer span of passages to each question. Thus, we did not conduct any experiments with these datasets in this study.

\paragraph{Distant supervision in IR for RC} Although SQuAD is a large-scale RC dataset, the domains and styles of its questions are limited. Incorporating distant supervision with supervised learning will be important for building open-domain QA systems. As distant supervision datasets, we can use the RC datasets without answer spans or other QA datasets without evidence passages, such as CuratedTREC~\cite{BaudisS15}, WebQuestions~\cite{BerantCFL13}, or WikiMoviews~\cite{MillerFDKBW16}.

Distant supervision in IR for RC has not yet been fully established. \citeauthor{WangAAAI2018} reported that their ranker trained with distant supervision  performed far worse than the oracle performance~\cite{WangAAAI2018}. 
In the future, we need to investigate the effects of using distant supervision in IR for RC, including the effect of adding adversarial distracting sentences~\cite{JiaL17}.

\paragraph{Neural IR} Neural ranking models are currently a popular topic in IR~\cite{MitraC17}. There are roughly two groups: embedding space models~\cite{HuangHGDAH13,ShenHGDM14b,NalisnickMCC16} and interaction-based models~\cite{
GuoFAC16, 
XiongDCLP17, 
HuiYBM17, 
DaiXC018, 
HuiYBM18}. 

In the embedding space models, the query and documents are first embedded into continuous vectors, and the ranking is calculated from their embeddings' similarity. These models are faster than interaction-based ones, and we can use them as initial rankers 
in order to alleviate the problem of lexical variation.

Most of the recent models use an interaction mechanism between the query and document words for accuracy.
These neural approaches give a relevance score to the whole document for training and tend to be data-hungry~\cite{MitraC17}. By contrast, our experimental results showed that our MTL approach accurately learns from a small amount of data.

\paragraph{Multi-task and transfer learning} MTL and transfer learning~\cite{PanY10} play a key role in building intelligent QA systems when the amount of available data is limited. \citeauthor{McCannBXS17} used contextualized word embeddings, called CoVe, trained in machine translation to improve the accuracy of RC~\cite{McCannBXS17}. \citeauthor{PetersNIGCLZ18} proposed contextualized word embeddings, called ELMo, trained in language modeling. Adding ELMo representations to existing models showed significant improvements on six challenging NLP problems including RC~\cite{PetersNIGCLZ18}. \citeauthor{Yu18} used a complementary data augmentation technique to enhance the training data of RC by using a translation model. The technique paraphrases examples by translating the original sentences from English to French and then back to English~\cite{Yu18}. These techniques can be used with our models.

\paragraph{Traditional IR-based QA systems} Most traditional systems focus on factoid questions, which can be answered with named entities, and have a pipeline architecture consisting of at least three components: question analysis, IR, and answer extraction~\cite{HirschmanG01}. The systems reformulate queries to enable their IR method to cover many textual variants. However, their reformulation is dependent on the redundancy of the knowledge source~\cite{BrillDB02}, and thus, they do not work well on smaller corpora. A deeper understanding of natural language is needed to overcome their limitations.

Moreover, there are two approaches to IR for QA: one is to index each passage as a separate document and retrieve them; the other one is to retrieve long {\it documents} from a corpus first and then find relevant short {\it passages} from the retrieved documents~\cite{TellexKLFM03}. Exploring the potential of such a two-stage IR in an end-to-end neural network model would be worthwhile. In particular, the work of \citeauthor{ChoiHUPLB17} \shortcite{ChoiHUPLB17} is related to the second stage of passage retrieval: it selects a few sentences from a long document (guaranteed to be relevant to a given question) and then generates the final answer from the selected sentences.

\section{Conclusion}
This study considered the task of \textit{machine reading at scale} (MRS) enabling QA based on a set of passages as a knowledge source. We improved IR for reading comprehension (RC).

Regarding the originality of our work, we believe our study makes two main contributions. First, we proposed the \textit{Retrieve-and-Read} model, which is based on a simple and effective approach that incorporates IR and RC tasks by using supervised multi-task learning (MTL). In the conventional reinforcement approach of joint learning of IR and RC tasks~\cite{WangAAAI2018}, the IR component is indirectly trained with a distant supervision reward based on RC predictions. Our model directly minimizes the joint loss of IR and RC in order that the IR component, which shares the hidden layers with the RC component, can be also trained with correct answer spans. Next, we created a new dataset, Jp-News, by using crowdsourcing in the same way as SQuAD. Jp-News is suitable for making evaluations of IR for RC tasks, because it consists of a set of news articles that contain similar passages on the same topics and a set of clear-intent long questions.

The key strength of this study is the high accuracy of our MRS system, particularly our IR component. While this study was limited to supervised learning, our MTL approach achieved state-of-the-art performance on a standard benchmark, in answering SQuAD questions using the full Wikipedia as the knowledge source. We also thoroughly evaluated the effectiveness of supervised MTL by investigating the individual contributions of our model components. The experimental results demonstrated the effectiveness of learning from answer spans in IR for RC. We believe that this finding will contribute to the development of MRS systems. Moreover, our approach can be easily applied to other state-of-the-art RC neural networks such as QANet~\cite{Yu18}. The existing RC methods could be extended into ones enabling QA from a corpus and handling questions that have no answer in the reading passage. Finally, the experimental results on our new dataset showed the capability of retrieving and reading passages in a non-English language without linguistic knowledge. 

Future work will involve exploring the potential of using distant supervision and enabling our model to combine disjoint pieces of textual evidence. 

\bibliographystyle{ACM-Reference-Format}
\bibliography{nishida-cikm18}

\appendix

\section{The Jp-News dataset}
\label{sec:dataset}
This section describes the Jp-News dataset consisting of QA data created by crowdworkers on a set of Japanese news articles.

\subsection{Dataset Collection}
The data collection consisted of three stages, as in the case of SQuAD~\cite{RajpurkarZLL16}.

\subsubsection{Passage curation} 
We crawled 14,804 articles and randomly sampled 5,000 articles, published from 17 June to 20 September 2017. We extracted individual paragraphs (passages) and stripped images and captions from each paragraph. The result was $12,485$ paragraphs. We partitioned the articles randomly into a training set ($80\%$), a development set ($10\%$), and a test set ($10\%$). 

\subsubsection{Question-answer collection}
We employed crowdworkers located in Japan to create questions. On each paragraph, three crowdworkers were tasked with asking and answering five questions on the content of that paragraph. The questions had to be entered in a text field, and the answer spans had to be selected in the paragraph. The workers were encouraged to ask questions in their own words in a way that other people could understand their questions without seeing the article. For example, "How old was he?" is a bad question despite the fact that "he" is uniquely determined in the article. "How old was the MVP in the MLB world series 2016?" is an example of a good question.

\subsubsection{Additional answers collection} 
We obtained two additional answers for each question. Each crowdworker was shown only the questions along with the paragraphs of an article and was asked to select the shortest span in the paragraph that answered the question. In total, we obtained 82,592 questions and 224,908 answers.

\subsection{Dataset Analysis}
We analyzed the dataset, from the viewpoint of articles, passages, questions, and answers, in order to demonstrate its characteristics in comparison with SQuAD. Table~\ref{tb:dataset_appendix} shows the statistics of each item in the Jp-News dataset. We used the JTAG tokenizer \cite{FuchiT98}.

\subsubsection{Articles}
The number of articles (5,000) is quite a bit larger than that of SQuAD (536). The articles cover a wide range of news categories: Local (26.1\%), World (20.0\%), Sports (13.7\%), Politics (13.0\%), Weather (12.0\%), Business (8.5\%), and Others (6.9\%).

Moreover, Jp-News contains a series of news articles that describe the same topic. This is in contrast to SQuAD, which was created from Wikipedia where most descriptions are expected to be stated only once. The crawled articles have hyperlinks to their related articles; 4.6\% of the articles in the development set have hyperlinks to other articles in the same set.

\subsubsection{Passages}
\begin{figure}[t!]
\centering
\includegraphics[width=.44\textwidth]{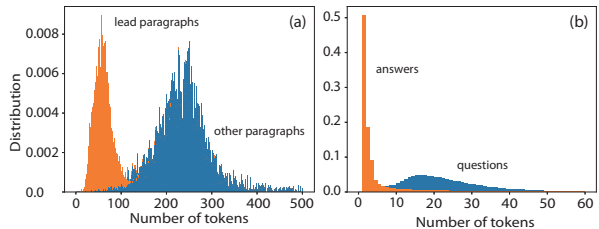}
\caption{Distribution of number of tokens. (a) Lead paragraphs and other paragraphs. (b) Answers and questions.}
\label{fig:tokens}
\end{figure}

The mean number of passage tokens (180.5) is slightly larger than that of SQuAD (140.3). The distribution of passage lengths consists of a mixture of two distributions: lead paragraphs, which summarize the main topic of articles, and other paragraphs, as shown in Figure~\ref{fig:tokens}. The existence of lead paragraphs and paragraphs of related articles makes it difficult for IR models to find the most relevant paragraph to each question.

\begin{table}[t!]
\centering
\caption{Number and mean length (in tokens) of each item in the Jp-News dataset. } 
\label{tb:dataset_appendix}
\begin{tabular}{lccc}\hline
& train & dev & test \\ \hline
Number of articles & 4,000 & 500 & 500 \\
Number of questions & 66,073 & 8,247 & 8,272 \\
Number of passages & 10,024 & 1,214 & 1,247 \\ 
Number of answers & 179,908 & 22,500 & 22,500 \\ 
Mean length of questions & 21.9 & 21.8 & 21.9 \\
Mean length of passages & 181.4 & 176.2 & 177.7 \\ 
Mean length of answers & 4.3 & 4.5 & 4.2 \\ 
\hline
\end{tabular}
\end{table}

\subsubsection{Questions}
The mean number of tokens is 21.9, and it is larger than that of SQuAD (11.4). Concrete questions are suitable for finding passages relevant to the questions in a corpus.

\subsubsection{Answers}
The mean number of tokens covering each answer string is 4.3, and that of SQuAD is 3.4. Table~\ref{tb:answer} shows the answer-type distributions of the Jp-News dataset. We can see that Jp-News contains a larger number of numeric, location, and clause answers than SQuAD does. 

\begin{table}[t!]
\caption{Answer type distributions on Jp-News and SQuAD. We manually examined 300 randomly sampled question-answer pairs of the Jp-News dataset.}
\label{tb:answer}
\centering
\begin{tabular}{lcc}
\hline
Answer type & SQuAD & Jp-News \\ \hline
Date & 8.9\% & 5.7\% \\
Other Numeric & 10.9\% & 21.7\% \\
Person & 12.9\% & 12.7\%\\
Location & 4.4\%& 19.7\%\\
Other Entity & 15.3\%& 9.0\%\\
Common Noun Phrase & 31.8\%& 18.7\%\\
Adjective Phrase & 3.9\%& 0.3\%\\
Verb Phrase & 5.5\%& 1.0\%\\
Clause & 3.7\%& 9.7\%\\
Other & 2.7\%& 1.7\%\\
\hline
\end{tabular}
\end{table}

\end{document}